\documentclass[final]{cvpr}

\usepackage{times}
\usepackage{epsfig}
\usepackage{graphicx}
\usepackage{amsmath}
\usepackage{amssymb}
\usepackage{cite,multirow,subfig}
\usepackage{lipsum}

\usepackage[pagebackref=true,breaklinks=true,colorlinks,bookmarks=false]{hyperref}

\newcommand\blfootnote[1]{%
\begingroup
\renewcommand\thefootnote{}\footnote{#1}%
\addtocounter{footnote}{-1}%
\endgroup
}

\begin{document}

\title{Fourier Contour Embedding for Arbitrary-Shaped Text Detection}

\author{Yiqin~Zhu$^{1\dag}$,
Jianyong~Chen$^{1\dag}$,
Lingyu~Liang$^{1,3*}$,
Zhanghui~Kuang$^{2*}$,
Lianwen~Jin$^{1,3}$,
Wayne Zhang$^{2,4,5}$
\\
$^1$South China University of Technology\ \ \
$^2$SenseTime Research\ \ \
$^3$Pazhou Lab\\
$^4$Qing Yuan Research Institute, Shanghai Jiao Tong University
$^5$Shanghai AI Laboratory, Shanghai, China
\\
{\tt \small
scut\_zhuyiqin@163.com,
\{theochan666,lianglysky,lianwen.jin\}@gmail.com
}\\
{\tt \small \{kuangzhanghui,wayne.zhang\}@sensetime.com}
}

\maketitle
\pagestyle{empty}
\thispagestyle{empty}

\begin{abstract}
One of the main challenges for arbitrary-shaped text detection is to design a good text instance representation that allows networks to learn diverse text geometry variances. Most of existing methods model text instances in image spatial domain via masks or contour point sequences in the Cartesian or the polar coordinate system. However, the mask representation might lead to expensive post-processing, while the point sequence one may have limited capability to model texts with highly-curved shapes. To tackle these problems, we model text instances in the Fourier domain and propose one novel Fourier Contour Embedding (FCE) method to represent arbitrary shaped text contours as compact signatures. We further construct FCENet with a backbone, feature pyramid networks (FPN) and a simple post-processing with the Inverse Fourier Transformation (IFT) and Non-Maximum Suppression (NMS). Different from previous methods, FCENet first predicts compact Fourier signatures of text instances, and  then reconstructs text contours via IFT and NMS during test. Extensive experiments demonstrate that FCE is accurate and robust to fit contours of scene texts even with highly-curved shapes, and also validate the effectiveness and the good generalization of FCENet for arbitrary-shaped text detection. Furthermore, experimental results show that our FCENet is superior to the state-of-the-art (SOTA) methods on CTW1500 and Total-Text, especially on challenging highly-curved text subset.
\end{abstract}
\blfootnote{$^\dag$Yiqin~Zhu and Jianyong~Chen contributed equally to this work.}
\blfootnote{$^*$\textbf{Corresponding authors: Lingyu~Liang, Zhanghui~Kuang.} This research is supported by NSFC (Grant No.61936003), GD-NSF (Grant No.2017A030312006, No.2019A1515011045), SenseTime Research Fund for Young Scholars and CAAI-Huawei MindSpore Open Fund.}


\section{Introduction}
Benefiting from the development of object detection~\cite{dai2017deformable, Lin_2017_CVPR,lin2017focal,redmon2016you,tian2019fcos} and instance segmentation~\cite{he2017mask,liu2018path}, text detection has achieved significant progress~\cite{zhou2017east,tian2016detecting,liao2018textboxes++,xu2019geometry,Long_2018_ECCV,tian2019learning,baek2019character,wang2019efficient,wang2019shape,zhang2019look,wang2020textray,liao2020real,zhang2020deep,wang2020contournet}.
 Text detection methods can be roughly divided into segmentation-based approaches~\cite{xie2019scene,tian2019learning,wang2019efficient,xu2019textfield,wang2019shape,Long_2018_ECCV,tang2019seglink++,tian2016detecting,shi2017detecting,ma2020relatext}, and regression-based approaches~\cite{zhang2019look,xue2019msr,zhang2020deep,wang2020textray,liu2020abcnet}.

\begin{figure}[t]
  \centering
  \begin{tabular}{cc}
  \multicolumn{2}{c}{\includegraphics[width=3in]{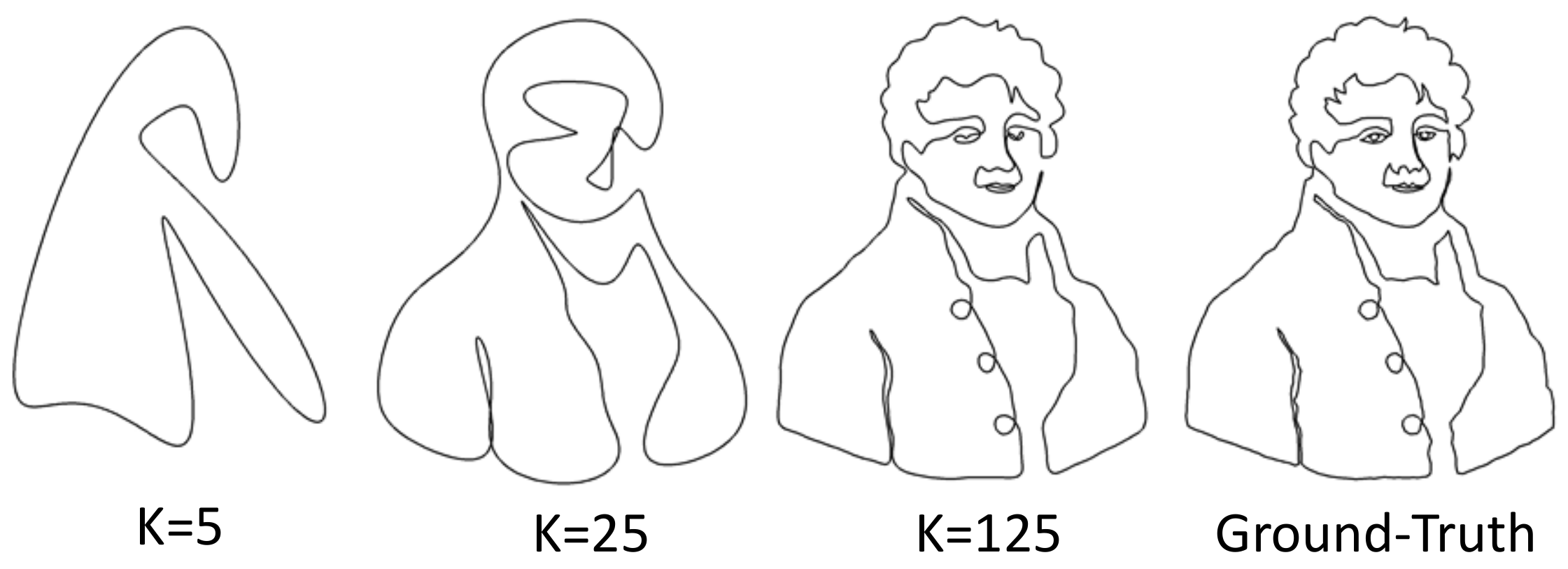}}\\
  \multicolumn{2}{c}{(a) Fourier contour fitting with progressive approximation.}\vspace{5pt}\\
  \includegraphics[height=0.95in]{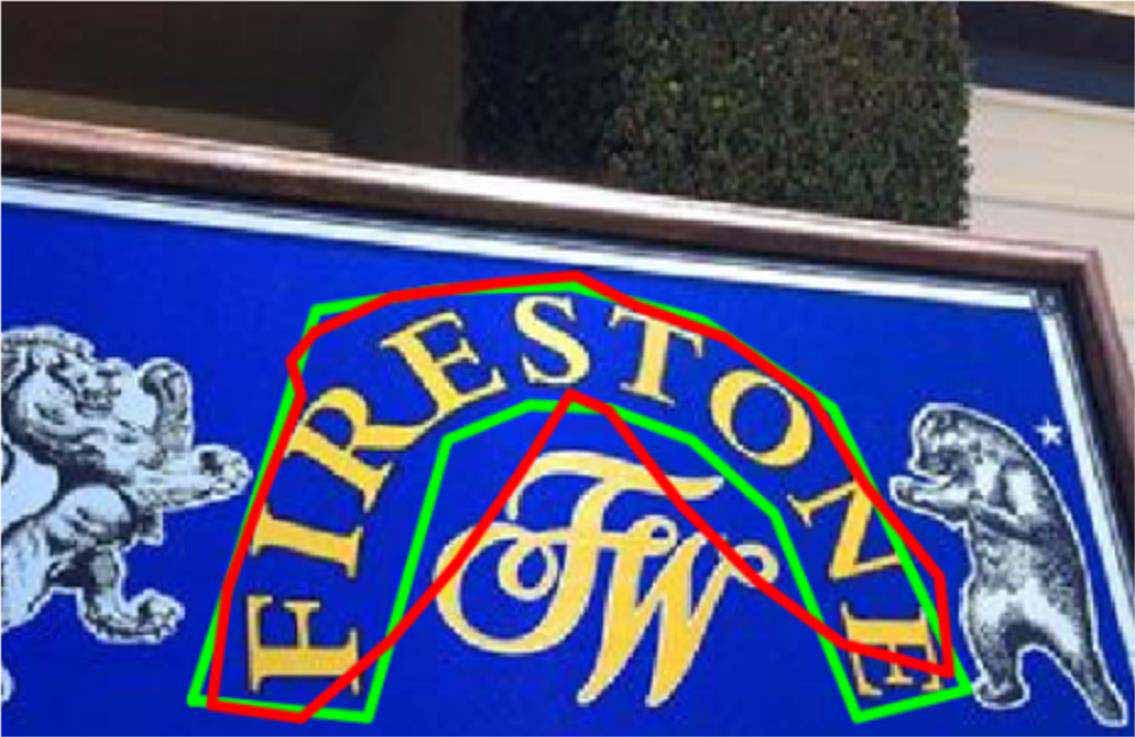}& \includegraphics[height=0.95in]{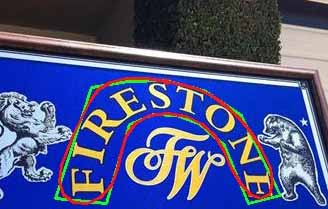}\\
  (b) TextRay contour~\cite{wang2020textray} & (c) Fourier contour \\
  \end{tabular}
  \caption{Comparison with Fourier contour and TextRay contour~\cite{wang2020textray} representations.
   (a) shows Fourier contour can fit extremely complicated object shapes and get better approximation as the Fourier degree $k$ increases. (b) and (c) compare the TextRay contours and our proposed Fourier contours, where the ground-truth contours are in green and the reconstructed ones are in red. TextRay fails to model highly-curved texts (best viewed in color). \label{fig_fourier}}
\end{figure}

Recent research focus
has shifted from horizontal or multi-oriented text detection~\cite{liao2018textboxes++,tian2016detecting,xu2019geometry,zhou2017east} to more challenging arbitrary-shaped
text detection~\cite{baek2019character,liao2020real,Long_2018_ECCV,tian2019learning,wang2020textray,wang2019shape,wang2019efficient,wang2020contournet,zhang2019look,zhang2020deep}. Compared to multi-oriented text detection, text instance representations play an indispensable role in arbitrary-shaped text detection.
A good representation should be simple and compact with good generalization ability to avoid overfitting. However, designing a compact text instance representation is not straightforward, because fitting diverse geometry variances of text instances is challenging. Existing arbitrary-shaped text detection approaches
represent text instances in the spatial domain of images.
They models texts via per-pixel masks~\cite{wang2020contournet,wang2019shape,wang2019efficient,baek2019character,Long_2018_ECCV,liao2020real,tian2019learning,xu2019textfield},
 contour point sequences in the Cartesian system~\cite{zhang2020deep,zhang2019look,liu2020abcnet} or those in the polar system~\cite{wang2020textray}.
Spatial domain based methods have clear drawbacks. Mask representation may lead to intrinsically computationally expensive post-processing, and frequently requires large training data. And contour point sequences may have limited capability to model highly-curved texts. 

In this paper, we model text instance contours in the Fourier domain instead of the spatial domain via
the Fourier transformation, which can fit any closed contour with progressive approximation in a robust and simple manner. Fig.~\ref{fig_fourier}a illustrates that Fourier transformation can accurately fit extremely complicated shapes (\eg, a portrait sketch) with very compact signatures (\eg, $K=125$ only), and  shows that as the Fourier degree $k$ increases, the reconstructed shape approximates the ground truth better.
Compared to TextRay~\cite{wang2020textray}, a SOTA text contour point sequence in the polar coordinate system,
 our proposed Fourier contour representation can model high-curved texts better as shown in
 Fig.~\ref{fig_fourier}b-c.

To this end, we propose Fourier Contour Embedding (FCE) method to convert text instance contours from point sequences into Fourier  signature vectors. Firstly, we propose a resampling scheme to obtain a fixed number of dense points on each text contour. To maintain the uniqueness of the resulted Fourier signature vector, we set the right-most intersection between the text contour and the horizontal line through the text center point as the sampling start point, fix the sampling direction as the clockwise direction, and keep the sampling interval along the text contour unchanged. Secondly, the sampled point sequences of contours in the spatial domain are embedded into the Fourier domain via the Fourier transformation (FT).

The advantages of FCE for text instance representation are three-fold:
\begin{itemize}
  \item \textbf{Flexible:} Any closed contour, including extremely complicated shapes, can accurately be fitted;
  \item \textbf{Compactness:} The Fourier signature vectors are compact of our method. In our experiments, Fourier degree $K=5$ can achieve very accurate approximation of texts.
  \item \textbf{Simplicity:} The conversion between a sampled point sequence and a Fourier signature vector of text contours is formulated as FT and Inverse FT. So the FCE method is easy to implement without introducing complex post-processing.
\end{itemize}

 Equipped with the FCE, we further construct FCENet for arbitrary-shaped text detection.
 Particularly, it consists of a backbone of ResNet50 with deformable convolutional networks (DCN)~\cite{zhu2019deformable}, feature pyramid networks (FPN)~\cite{Lin_2017_CVPR} and the Fourier prediction header. The header has two individual branches. Namely, the classification branch, and the regression branch. The former predicts text region masks and text center region masks. The latter predicts
 text Fourier signature vectors in the Fourier domain, which are fed into the Inverse Fourier Transformation (IFT) to reconstruct text contour point sequences. Ground truth text contour point sequences are used as supervision signals.
 Thanks to the resampling scheme of FCE, our loss in the regression branch is compatible across different datasets, although datasets such as CTW1500~\cite{liu2019curved} and Total-Text~\cite{ch2017total} have different numbers of contour points for each text instance.

Experiments validate the effectiveness and good generalization ability of FCENet for arbitrary shaped text detection. Moreover, our FCENet is superior to the state-of-the-art (SOTA) methods on CTW1500 and Total-Text, especially on their highly-curved text subset.

We summarize the contributions of this work as follows:
\begin{itemize}
  \item We propose {Fourier Contour Embedding (FCE)} method, which can accurately approximate
   any closed shapes, including arbitrary shaped text contours, as compact Fourier signature vectors.
  \item We propose {FCENet} which first predicts Fourier signature vectors of text instances in the Fourier domain, and then reconstructs text contour point sequences in the image spatial domain via Inverse Fourier Transformation (IFT). It can be learned end-to-end, and be inferred without any complex post processing.
  \item We extensively evaluate the proposed FCE and FCENet. Experimental results validate the good representation of FCE, especially on highly-curved texts, the generalization ability of FCENet when training on small datasets. Moreover, it has been shown that FCENet achieves the state-of-the-art performance on CTW1500 and Total-Text.
\end{itemize}



\section{Related Work}
\subsection{Segmentation-Based Methods}
These methods mainly draw inspiration from semantic
segmentation, which implicitly encodes text instances with per-pixel masks~\cite{xie2019scene,tian2019learning,wang2019efficient,xu2019textfield,wang2019shape,Long_2018_ECCV,tian2016detecting,shi2017detecting,baek2019character,yao2016scene}.
Most of these methods follow a component-grouping paradigm, which first detect components of scene text instances  and then aggregate these components to obtain final mask outputs.

For pixel-based methods, pixel-level score maps are firstly obtained using instance/semantic segmentation framework, and then text pixels are grouped to obtain the output text masks~\cite{xie2019scene,tian2019learning,wang2019efficient,xu2019textfield}. To further improve the performance, some methods would perform prediction on a transformed space, and then reconstruct the final maps. For example, Tian~\etal~\cite{tian2019learning} assumed each text instance as a cluster and predicted an embedding map via pixel clustering; TextField~\cite{xu2019textfield} generates candidate text parts via linking neighbor pixels with a deep direction field.

For segment-based methods, segments containing a part of a word or text line (fragments)~\cite{wang2019shape,tang2019seglink++,Long_2018_ECCV,tian2016detecting,shi2017detecting,ma2020relatext} or characters~\cite{baek2019character,yao2016scene} are firstly detected, and then segments are grouped into the whole words/text-line. PSENet~\cite{wang2019shape} detects each text instance with corresponding kernels, and adopts a progressive scale algorithm to gradually expand the predefined kernels obtain the final detection. SegLink++~\cite{tang2019seglink++} achieves dense and arbitrary-shaped scene text detection using instance-aware component grouping with minimum spanning tree. CRAFT~\cite{baek2019character} obtains character-level detection and estimates the affinity between characters to achieve the final detection.

Some methods train the predictor in a transformed space, and reconstruct the output masks via the predicted features. For example, Tian~\etal~\cite{tian2019learning} constructed a discriminative representation via embedding pixels into a space where pixels of the same text tend to be in the same clusters and vise versa; Xu~\etal~\cite{xu2019textfield} proposed TextField to learn one direction field to separate adjacent text instances.


\subsection{Regression-Based Methods}
Regression-based methods are complementary to segmentation-based methods, which explicitly encodes text instances with contours (point sequences) of text regions. They aim to adopt the direct shape modeling of text instances to handle complex geometric variances~\cite{zhang2019look,zhang2020deep,wang2020textray,xue2019msr,zhou2017east,liao2018textboxes++}, and are often simpler and easier to train. However, the constrained representation capability of point sequences for complex text instances may limit the performance of the networks.

To tackle this problem, many modules are elaborately designed to further improve the flexibility of point sequence representation. LOMO~\cite{zhang2019look} introduces an iterative refinement module (IRM) and a shape expression module (SEM) to progressively refine the text localization of a direct regression. Zhang~\etal~\cite{zhang2020deep} used CNN to regress the geometry attributes (\eg, height, width, and orientation) of a series of small rectangular components divided from text instances, and introduced Graph Convolutional Network (GCN) to infer the linkages between different text components. TextRay~\cite{wang2020textray} formulates the text contours in the polar system and propose a single-shot anchor-free framework to learn the geometric parameters. Liu~\etal~\cite{liu2020abcnet} introduced Bezier curves to parameterize curved texts and achieved the SOTA performance in scene text spotting with BezierAlign.

Recent works indicate that effective contour modeling is essential for irregular text instances detection~\cite{zhang2019look,wang2020textray,zhang2020deep} and the downstream recognition~\cite{liu2020abcnet}. Therefore, it would be significant to design a flexible yet simple representation for arbitrary shaped text detection.

\begin{figure}[t]
  \centering
  \includegraphics[width=3in]{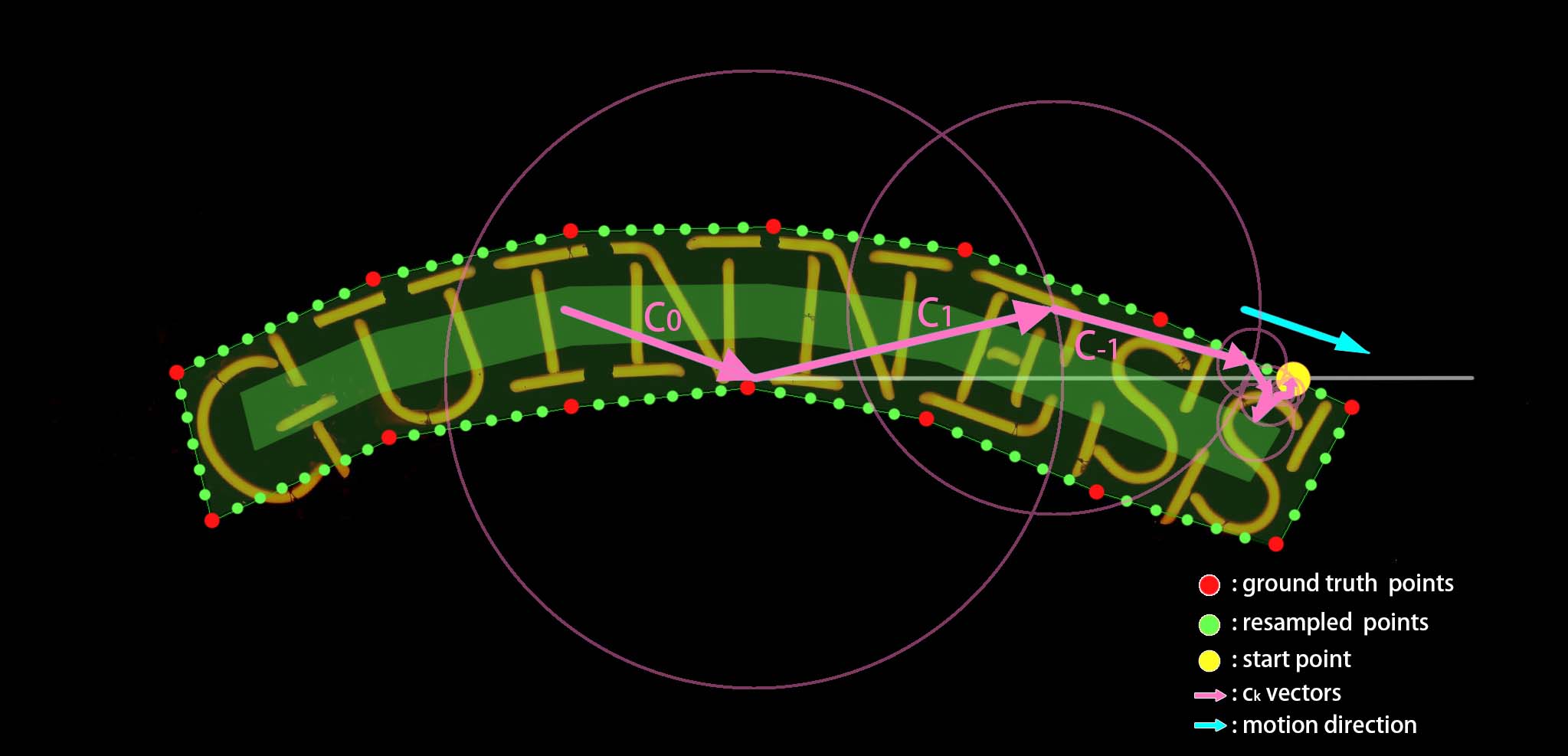}\\
  \caption{\small{Illustration of FCE. It contains two stages, where \textbf{Resampling} obtains dense point sequences (in green) based on ground truth points (in red); \textbf{Fourier Transformation} is used to compute Fourier coefficients $\textbf{c}_{k}$ with the resampled point sequences. A contour can be reconstructed by the combination of different fixed-frequency circular motions (indicated by pink circles) with the hand direction  $\textbf{c}_{k}$.}}\label{fig_resample}
\end{figure}

\subsection{Explicit vs. Implicit Text Shape Representation}
From a perspective of text shape representation, current model can be roughly divided into two categories. approaches which implicitly model text shapes via
per-pixel masks~\cite{wang2020contournet,wang2019shape,wang2019efficient,baek2019character,Long_2018_ECCV,liao2020real} or masks reconstructed by transformed features~\cite{tian2019learning,xu2019textfield},  and approaches which explicitly model text shapes using point sequences of contours in the Cartesian system~\cite{zhang2020deep,zhang2019look} or the polar system~\cite{wang2020textray}.

However, per-pixel masks may cause intrinsically high computational complexity of networks (\eg, complex post-processing) and require large training data, while point sequences sampled on contours may has limited representation capability and requires deliberately-designed refinement or inference~\cite{zhang2020deep,zhang2019look,wang2020textray,ma2020relatext}.

To tackle this problem, Liu~\etal~\cite{liu2020abcnet} introduced Bezier curves to parameterize curved texts, but the control point setting of Bezier curves may limit its representation capability for some cases, as shown in Sec.~\ref{sec_exp_chall}. In this paper, text instances are formulated in the Fourier domain, which allows to fit any closed continuous contour with robust and simple manners. In the following section, we would explore the potential of FCE for arbitrary shaped text detection.

\section{Approach}
In this section, we first introduce the proposed Fourier Contour Embedding (FCE) method, which can
approximate arbitrary-shaped text contours as compact Fourier signature vectors.
Then we propose FCENet to detect arbitrary-shaped texts, equipped with FCE.


\subsection{Fourier Contour Embedding}

We use one complex-value function $f:\mathbb{R}\mapsto\mathbb{C}$ of a real variable $t\in [0,1]$ to represent any text closed contour as follows:
\begin{equation}
f(t)=x(t)+iy(t),
\end{equation}
where $i$ represents the imaginary unit. $(x(t),y(t))$ denotes the spatial coordinate at the specific time $t$. Since $f$ is a closed contour, $f(t)=f(t+1)$. $f(t)$ can be reformulated by Inverse Fourier Transformation (IFT) as:
\begin{equation}
f(t)=f(t,\textbf{c})=\sum_{k=-\infty}^{+\infty}\textbf{c}_{k}e^{2\pi ikt}, \label{equ3}
\end{equation}
where $k \in \mathbb{Z}$ represents the frequency, and $\textbf{c}_{k}$ is the complex-value Fourier coefficient used to characterize the initial state of the frequency $k$. Each component $\textbf{c}_{k}e^{2\pi ikt}$
in Eq.~\ref{equ3} indicates a circular motion with fixed-frequency $k$ with a given initial hand direction vector $\textbf{c}_{k}$, Thus, the contour can be regarded as the combination of different frequent circular motions as the pink circles shown in Fig.~\ref{fig_resample}.
From Eq.~\ref{equ3}, we observe that the low frequency components are in charge of the rough text contours,
while the high are in charge of the details of contours. We empirically find that preserving $K-$lowest ($K=5$ in our experiments) frequencies only while discarding others can obtain satisfactory approximation of text contours, as shown in Fig.~\ref{fig_fourierK}.

Since we cannot obtain the analytical form of text contour function $f$ in real applications, we can
discretize the continual function $f$ into $N$ points as $\{f(\frac{n}{N})\}$ with $n\in [1,\dots, N]$. In this case, the $\mathbf{c}_k$ in Eq.~\ref{equ3} can be computed via the Fourier Transformation as:
\begin{equation}
\textbf{c}_{k}=\frac{1}{N}\sum_{n=1}^{N}f(\frac{n}{N})e^{-2\pi ik\frac{n}{N}}\label{4},
\end{equation}
where $\textbf{c}_{k}=u_k+iv_k$ with $u_k$ as the real part and $v_k$ as the image part of a complex number. Specially, when $k=0$, $\textbf{c}_{0}=u_{0}+iv_{0}=\frac{1}{N}\sum_n f(\frac{n}{N})$ is the center position of the contour. For any text contour $f$, our proposed Fourier Contour Embedding (FCE) method can represent it in the Fourier domain as a compact $2(2K+1)$ dimensional vector $[u_{-K},v_{-K},\cdots,u_0,v_0,\cdots,u_K,v_K]$, dubbed Fourier signature vector.

Our FCE method consists of two stages. Namely, the resampling stage and the Fourier transformation stage.
Concretely, in the resampling stage, we sample equidistantly a fixed number $N$ ($N=400$ in our experiments) points on the text contour, obtaining the resampled point sequence $\{f(\frac{1}{N}),\cdots,f(1)\}$.
Note that this resampling is necessary since different datasets have different numbers of ground truth points for text instances, and they are relatively small. \eg, there are $14$ in CTW1500~\cite{liu2019curved} while $4\sim 8$ in Total-Text~\cite{ch2017total}.
The resampling strategy makes our FCE is compatible to all datasets with the same setting.
In the Fourier transformation stages, the resampled point sequence is transformed into
its corresponding Fourier signature vector.



\textbf{Uniqueness of Fourier Signature Vector.}
From the above procedure of FCE, it is easy to see that different resampled point sequences can result in
different Fourier signature vectors even for the same text contour. To make the signature vector of one specific text unique, and more stable network training, we make constrains on the starting point, the sampling direction, and moving speed of $f(t)$:
\begin{itemize}
    \item \textbf{Starting point:} We set our starting point $f(0)$ (or $f(1)$) to be right most intersection point between the horizontal line through the center point $(u_{0},v_{0})$ and the text contour.
    \item \textbf{Sampling direction:} We always resample the points along the text contour in the clockwise direction.
    \item \textbf{Uniform speed:} We resample points uniformly on the text contour, and the distance between every two adjacent points keeps unchanged to ensure a uniform speed.
\end{itemize}



\begin{figure*}[t]
  \centering
  \includegraphics[width=7in]{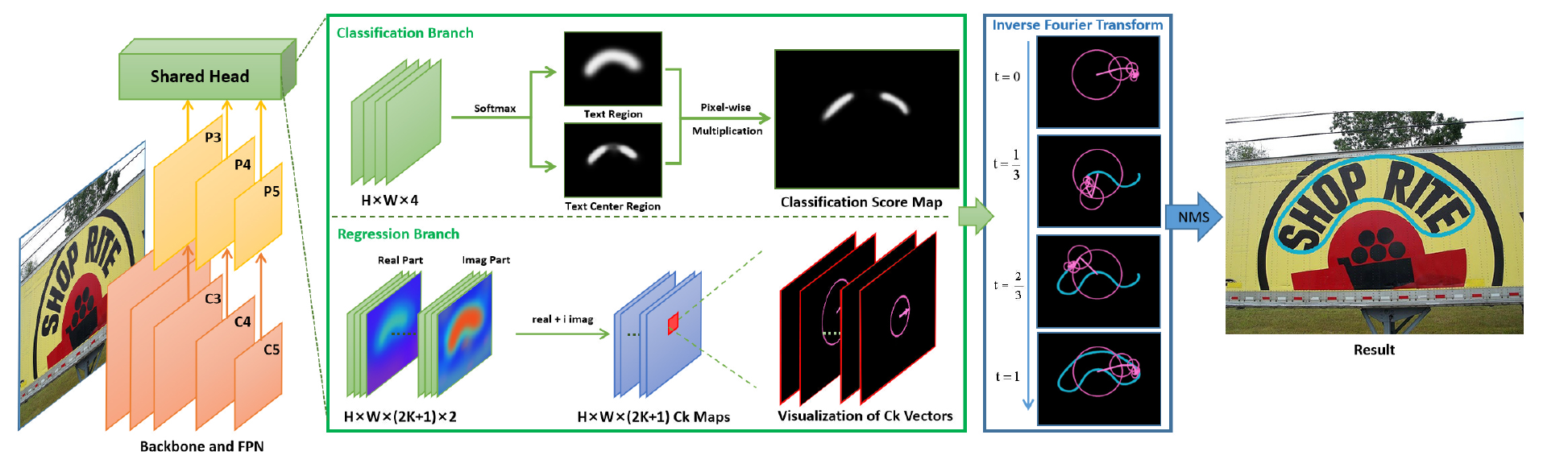}\\
  \caption{The overall framework of the proposed FCENet. Given an image, its features extracted by the backbone, and FPN, are fed into the shared header to detect texts. In the header,
    the classification branch predicts both the heat maps of text regions and those of text center regions, which are pixel-wise multiplied, resulting in the the classification score map. The regression branch
    predicts the Fourier signature vectors, which are used to reconstruct text contours via the Inverse Fourier transformation (IFT). Given the reconstructed text contours with corresponding classification scores, the final detected texts are obtained with non-maximum suppression (NMS)~\cite{1671883}. }\label{fig_framework}
\end{figure*}

\subsection{FCENet}
Equipped with the FCE, we further propose the anchor free network FCENet for arbitrary-shaped text detection.

\textbf{Network Architectures.}
Our proposed FCENet employs a top-down scheme.
As shown in Fig.~\ref{fig_framework},
it contains ResNet50~\cite{He_2016_CVPR} with DCN~\cite{zhu2019deformable} as backbone, and  FPN~\cite{Lin_2017_CVPR} as neck to extract multi-scale features, and the Fourier prediction header. We conduct prediction on the feature map P3, P4 and P5 of FPN.
The header has two branches, which are responsible for classification and regression respectively. Each branch consists of three $3\times3$ convolutional layers£¬ and one $1\times1$ convolutional layer, each of which is followed by one ReLU nonlinear activation layer.

In the classification branch,
we predict the per-pixel masks of Text Regions (TR). We find that Text Center Region (TCR) prediction
can further improve the performance. We believe this is because it can effectively filter out low-quality predictions around text boundaries.

In the regression branch, the Fourier signature vector of one text is regressed for each pixel in the text. To deal with text instances of different scales,  the features of P3, P4 and P5 are responsible for small, medium and large text instances, respectively. 

\begin{figure}[t]
  \centering
  \begin{tabular}{cc}
  \includegraphics[height=0.9in]{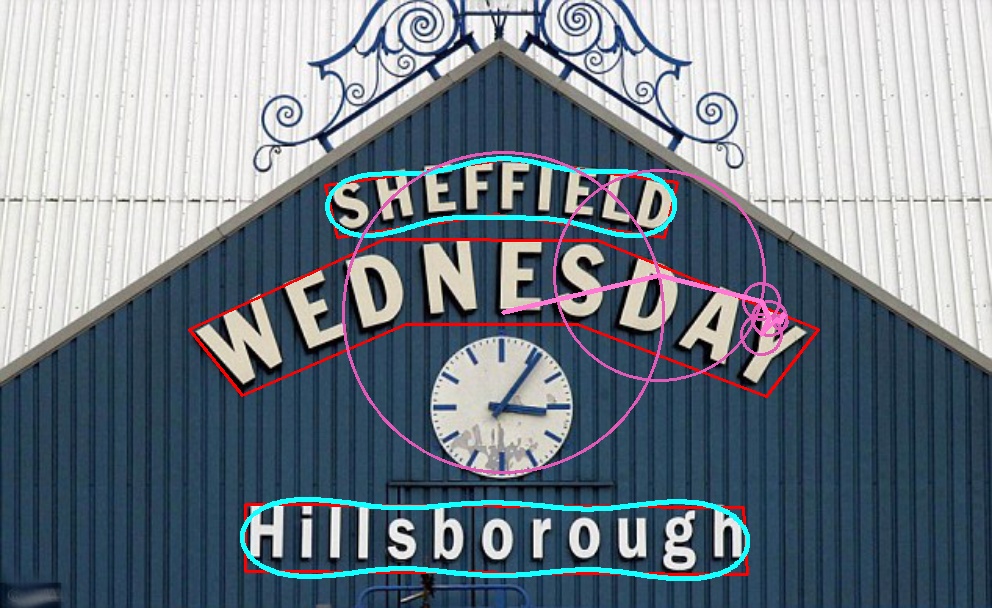}    & \includegraphics[height=0.9in]{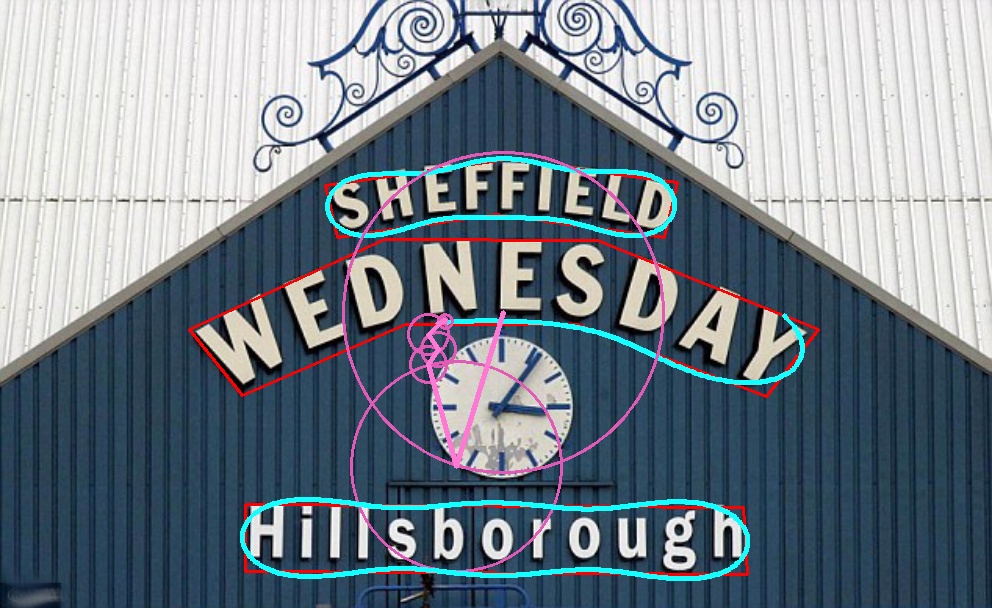}\\
  (a) $t=0$                          & (b) $t=1/3$ \vspace{5pt}\\
  \includegraphics[height=0.9in]{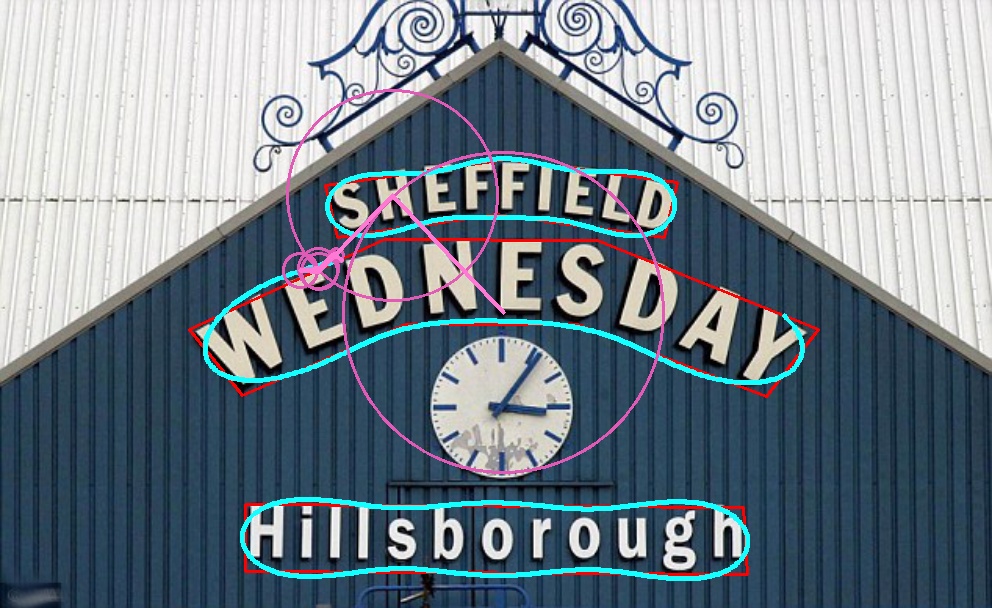}    & \includegraphics[height=0.9in]{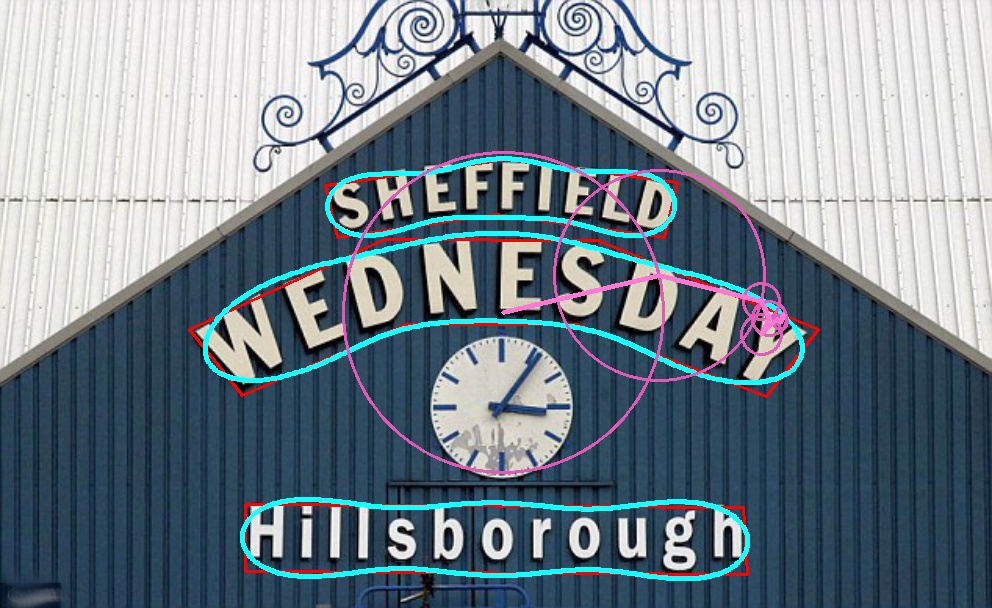}\\
  (c) $t=2/3$                        & (d) $t=1$ \\
  \end{tabular}
  \caption{Fourier contour reconstruction (blue) via IFT and NMS for arbitrary-shaped texts (red denotes ground-truth) at different time $t$.}\label{fig_IFT}
\end{figure}

The detection results would be reconstructed from the Fourier domain to the spatial domain by IFT and NMS, as shown in Fig.~\ref{fig_IFT}.




\textbf{Ground-Truth Generation.}
For the classification task, we use the method of~\cite{Long_2018_ECCV} to obtain text center region (TCR) masks via shrinking texts with the shrinking factor being 0.3 (see the green mask in Fig.~\ref{fig_resample} ).
For the regression task, we compute the Fourier signature vectors $\mathbf{\overline c}$ of the ground truth text contours via the proposed FCE method. Note that for all pixels in the mask of one text instance, we predict the text contour, and thus need one Fourier signature vector $\mathbf{\overline c}$ with the pixel
 being the $(0,0)$ point of the complex coordinate system.
 Different pixels in the same text instance share the same Fourier signature vector except $\textbf{c}_{0}$.

\textbf{Losses.} The optimization objective of FCE-base network is given by:
\begin{equation}
\mathcal{L} = \mathcal{L}_{cls}+\lambda \mathcal{L}_{reg},
\end{equation}
where $\mathcal{L}_{cls}$ and and $\mathcal{L}_{reg}$ are the loss for the classification branch and that for the regression branch, respectively. $\lambda$ is a parameter to balance $\mathcal{L}_{cls}$ and $\mathcal{L}_{reg}$. We fix $\lambda=1$ in our experiments. $\mathcal{L}_{cls}$ consists of two parts as:
\begin{equation}
\mathcal{L}_{cls} = \mathcal{L}_{tr}+\mathcal{L}_{tcr},
\end{equation}
where $\mathcal{L}_{tr}$ and $\mathcal{L}_{tcr}$ are the cross entropy loss for the text region (TR) and that of the text center region (TCR), respectively. To solve the sample imbalance problem, OHEM~\cite{Shrivastava_2016_CVPR} is adopted for $\mathcal{L}_{tr}$ with the ratio between negative and positive samples being $3:1$.

For $\mathcal{L}_{reg}$,
we do not minimize the distances between the predicted Fourier signature vectors and their corresponding
ground truth. In contrast, we minimize their reconstructed text contours in the image spatial domain which are more related to the text detection quality. Formally,
\begin{equation}
\mathcal{L}_{reg}=\frac{1}{N^{'}}\sum_{i\in \mathcal{T}}\sum_{n=1}^{N^{'}}w_i{{l_{1}}}(F^{-1}(\frac{n}{N^{'}},\mathbf{\overline c}_i),F^{-1}(\frac{n}{N^{'}},\hat{\mathbf{c}_i})),
\label{eq-reloss}
\end{equation}
where $l_1$ is the smooth$-l_1$ loss~\cite{NIPS2015_5638} used for regression, and
$F^{-1}(\cdot)$ is the IFT of Eq.~\ref{equ3}. $\mathcal{T}$ is the text region pixel index set.
$\mathbf{\overline c}_i$ and $\mathbf{\hat c}_i$ are the text ground truth Fourier signature vector and the predicted one for pixel $i$. $w_i=1$ if pixel $i$ in its corresponding text center region while 0.5 if not.
$N^{'}$ is the sampling number on the text contours. If $N^{'}$ is too small (typically $N^{'}<30$), it would probably cause over-fitting. Therefore, we fix $N^{'}=50$ in our experiments.

The regression loss is extremely important in our FCENet. In ablation studies of Sec.~\ref{sec_fcenent}, results show that it brings  absolute $6.9\%$ and $9.3\%$ h-mean improvement on CTW 1500 and Total-text respectively.


\section{Experiments}
In this section, we first verified the effectiveness of FCE to model text instances, compared with two recent SOTA arbitrary shaped text representation methods, \ie, TextRay~\cite{wang2020textray} and ABCNet~\cite{liu2020abcnet}.
We then evaluated FCENet for text detection.
Particularly, we conducted ablation studies for the effectiveness of each component, and the generalization ability by decreasing the training data; we also made extensive comparison with the recent SOTA methods on CTW1500~\cite{liu2019curved} and Total-Text~\cite{ch2017total} benchmarks. Since these benchmark datasets also contain a large amount of non-curved texts, we built a much more challenging subset containing highly-curved or highly irregular text for further evaluation.

\subsection{Datasets}
\textbf{CTW1500}~\cite{liu2019curved} contains both English and Chinese texts with text-line level annotations, where $1000$ images for training, and 500 images for testing.

\textbf{Total-Text}~\cite{ch2017total} was collected form various scenes, including text-like background clutter and  low-contrast texts, with word-level polygon annotations, where 1255 images for training and 300 images for testing.

\textbf{ICDAR2015}~\cite{7333942} is a multi-orientated and street-viewed dataset which consists of 1000 training and 500 testing images. The annotations are word-level with four vertices.

\subsection{Implementation Details}\label{sec_para}
The backbone of FCENet contains ResNet50 with DCN~\cite{zhu2019deformable} and a FPN~\cite{Lin_2017_CVPR}, as shown in Fig.~\ref{fig_framework}. Each of the regression and classification branch consists of three $3\times3$ convolutional layers and one $1\times1$ convolutional layer, whose kernel numbers are set as $[128, 64, 32, 32]$.  The text scale ranges  of P3, P4 and P5 are set to $[0, 0.4]$, $[0.3, 0.7]$, and $[0.6, 1]$ of the image size respectively, where the overlapping range is to increase the recall rate.

We resize images to $800\times800$ , and  adopt data augmentation strategies, including random crop, random rotations, random horizontal flipping, color jitter and contrast jitter during training. The models are trained using two 2080Ti GPUs with batch size set to 8. Stochastic gradient descent(SGD) is adopted as optimizer with the weight decay of 0.001, and the momentum of 0.9. The initialized learning rate is 0.001, which is reduced $0.8\times$ every 200 epoches.

During testing, the images are resized as follows: For CTW1500, we first resize the short edge of images to $640$, and then resize the long edge of the resulted images to $1280$ if it is bigger than $1280$. For Total-Text, we first resize the short edge of images to $960$, and then resize the long edge of the resulted images to $1280$ if it is bigger than $1280$. For ICDAR2015, we resize the long edge to $2020$ while keeping its original direction.

Previous methods in comparisons were implemented with their open source codes, and some of them were tested on MindSpore platform\footnote{https://github.com/mindspore-ai/mindspore}.

\begin{figure}[t]
  \centering
  \includegraphics[width=1in]{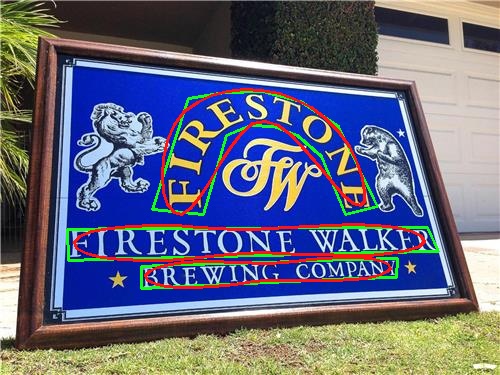}\hspace{.5pt}
  \includegraphics[width=1in]{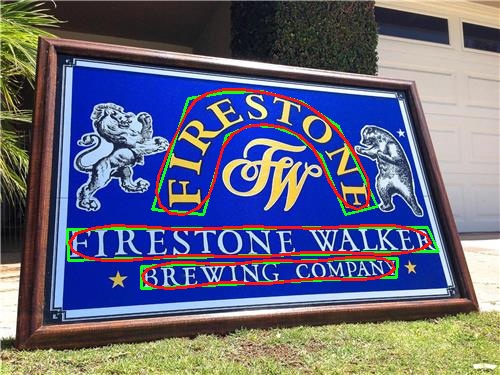}\hspace{.5pt}
  \includegraphics[width=1in]{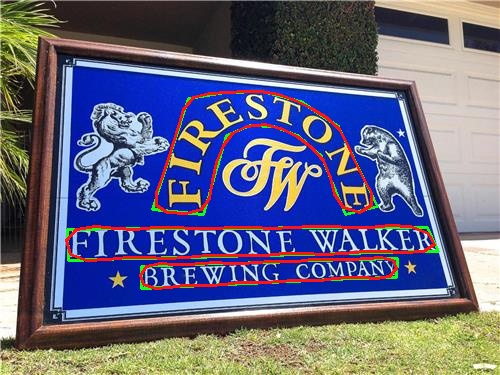}\vspace{2pt}\\

  \includegraphics[width=1in]{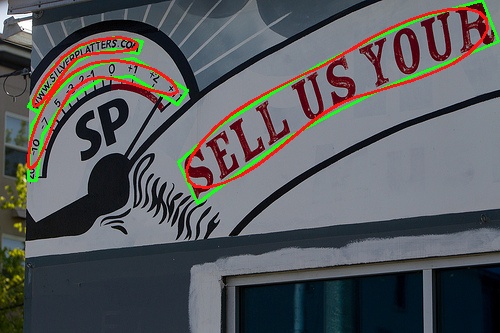}\hspace{.5pt}
  \includegraphics[width=1in]{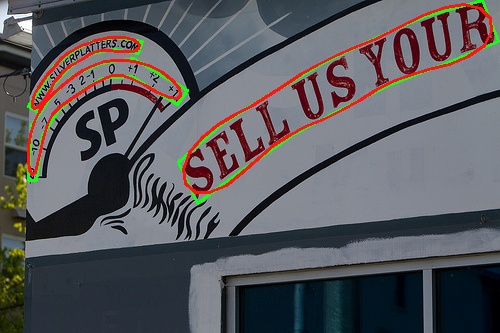}\hspace{.5pt}
  \includegraphics[width=1in]{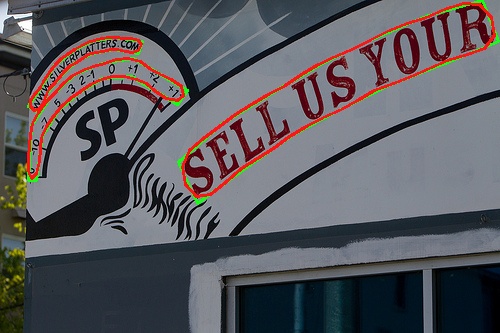}\vspace{-8pt}\\

  \subfloat[$K=3$]{\includegraphics[width=1in]{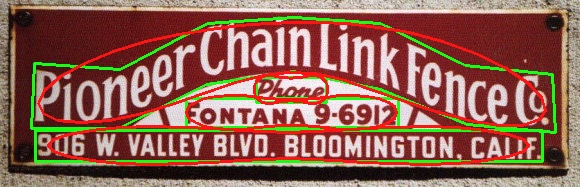}}\hspace{.5pt}
  \subfloat[$K=5$]{\includegraphics[width=1in]{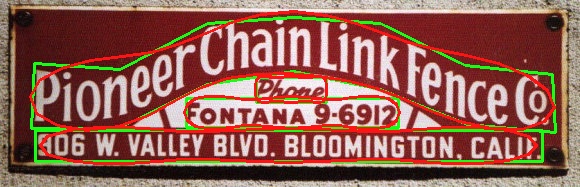}}\hspace{.5pt}
  \subfloat[$K=10$]{\includegraphics[width=1in]{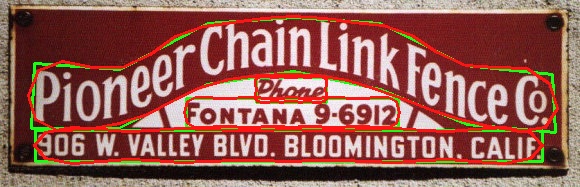}}
  \caption{Fourier contour fitting for arbitrary-shaped texts with increasing Fourier degree $K$, where green contours denotes ground-truth; red contours denotes FCE fitting results.}\label{fig_fourierK}
\end{figure}


\begin{figure}[t]
  \centering
  \includegraphics[width=1.5in]{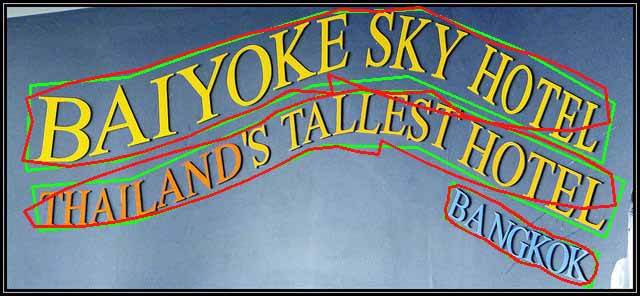}\hspace{.5pt}
  \includegraphics[width=1.5in]{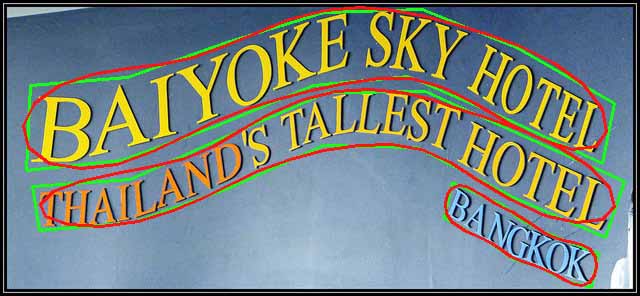}\vspace{-7pt}\\




  \subfloat[TextRay~\cite{wang2020textray}]{\includegraphics[width=1.5in]{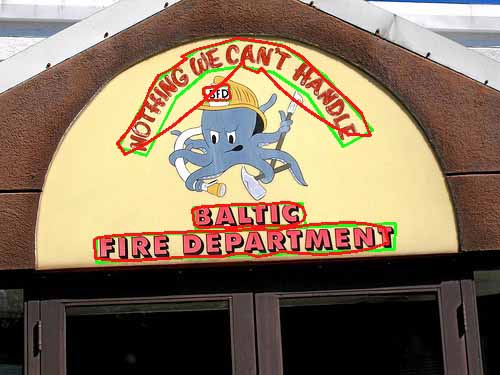}}\hspace{.5pt}
  \subfloat[Ours]{\includegraphics[width=1.5in]{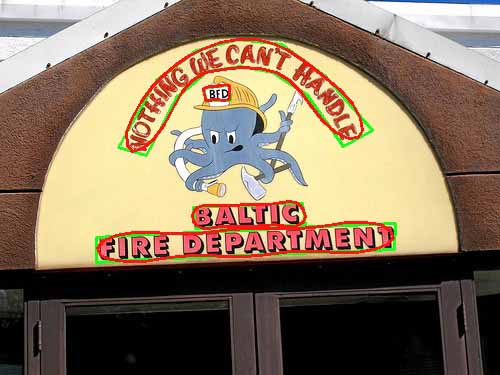}}

 \caption{Comparisons between different text representations in terms of contour fitting ability. Green contours denote ground-truth, while red ones denote the fitting results. }\label{fig_fit-vs}
\end{figure}

\begin{table*}[t]
\centering
\small{
\caption{Comparison with related methods on CTW1500, Total-Text and ICDAR2015, where 'Ext.' denotes extra training data and 'FCENet\dag' denotes FCENet using ResNet50 as a backbone without DCN.}
\begin{tabular}{|r|c|c|c|c|c||c|c|c||c|c|c|}
  \hline
      \multirow{2}{*}{Methods}& \multirow{2}{*}{Paper}& \multirow{2}{*}{Ext.}&
      \multicolumn{3}{c||}{CTW1500}&\multicolumn{3}{c||}{Total-Text}&
      \multicolumn{3}{c|}{ICDAR2015}\cr\cline{4-12}
      &&&R($\%$)      & P($\%$)      & F($\%$) &R($\%$)      & P($\%$)      & F($\%$) &R($\%$)      & P($\%$)      & F($\%$)\\
  \hline
  \hline
  TextSnake~\cite{Long_2018_ECCV}           &ECCV'18&$\surd$    &\textbf{85.3}	      &67.9	          &75.6      &74.5	&82.7	&78.4   &80.4 &84.9 &82.6\\
  \hline
  SegLink++~\cite{tang2019seglink++}        &PR'19&$\surd$      &79.8	      &82.8	          &81.3   &80.9  &82.1   &81.5   &80.3  &83.7  &82.0\\
  \hline
  SAEmbed~\cite{tian2019learning}           &CVPR'19&$\surd$    &77.8	      &82.7	          &80.1   &-&-&-   &85.0  &88.3  &86.6\\
  \hline
  CRAFT~\cite{baek2019character}            &CVPR'19&$\surd$    &81.1	      &86.0	          &83.5   &79.9	&87.6	&83.6   &84.3   &89.8  &86.9\\
  \hline
  PAN~\cite{wang2019efficient}              &ICCV'19&$\times$          &77.7	      &84.6	          &81.0   &79.4	&88.0	&83.5   & 77.8  &82.9  &80.3\\
  \hline
  PAN~\cite{wang2019efficient}              &ICCV'19&$\surd$    &81.2	      &86.4	          &83.7   &81.0	&\textbf{89.3}	&85.0   &81.9  &84.0  &82.9\\
  \hline
  PSENet~\cite{wang2019shape}               &CVPR'19&$\times$          &75.6	      &80.6	          &78.0   &75.1  &81.8  &78.3   &79.7  &81.5  &80.6\\
  \hline
  PSENet~\cite{wang2019shape}               &CVPR'19&$\surd$    &79.7	      &84.8	          &82.2   &84.0  &78.0  &80.9   &84.5  &86.9  &  85.7\\
  \hline
  LOMO~\cite{zhang2019look}                 &CVPR'19&$\surd$    &76.5	      &85.7	          &80.8   &79.3	&87.6	&83.3   &83.5  &91.3  &87.2\\
  \hline
  DB~\cite{liao2020real}                    &AAAI'20&$\surd$    &80.2	      &86.9	          &83.4   &82.5	&87.1	&84.7   &83.2  &\textbf{91.8}  &\textbf{87.3}\\
  \hline
  Boundary~\cite{wang2020boundary}          &AAAI'20&$\surd$    &-&-&-        &83.5         &85.2           &84.3 &88.1   &82.2   &85.0     \\
  \hline
  DRRG~\cite{zhang2020deep}                 &CVPR'20&$\surd$    &83.0	      &85.9	          &84.5   &\textbf{84.9}	&86.5	&85.7   &84.7  &88.5  &86.6  \\
  \hline
  ContourNet~\cite{wang2020contournet}      &CVPR'20&$\times$          &84.1         &83.7	          &83.9   &83.9	&86.9	&85.4     &\textbf{86.1}  &87.6  &86.9\\
  \hline
  TextRay~\cite{wang2020textray}            &MM'20&$\surd$    &80.4	          &82.8	          &81.6   &77.9	&83.5	&80.6   &-&-&-\\
  \hline
  ABCNet~\cite{liu2020abcnet}               &CVPR'20&$\surd$    &78.5         &84.4           &81.4   &81.3  &87.9   &84.5     &-&-&-\\
  \hline
  \hline
  \textbf{FCENet\dag}                       &Ours&$\times$          &80.7        &85.7  &83.1 &79.8	&87.4	&83.4   &84,2  &85.1  &84.6  \\
  \hline
  \textbf{FCENet}                           &Ours&$\times$          &83.4         &\textbf{87.6}  &\textbf{85.5} &82.5	&\textbf{89.3}	&\textbf{85.8}   &82.6  &90.1  &86.2  \\
  \hline
\end{tabular}\label{table3}}
\end{table*}

\subsection{Evaluation of FCE}
\textbf{Basic Evaluation.} Theoretically, any closed continuous contour can be fitted by Fourier contour with a better approximation via increasing Fourier degree $K$ of FCE. Results of Fig.~\ref{fig_fourierK} indicates that only small $K$ can obtain satisfactory fitting for most of arbitrary-shaped texts, which verifies the strong representation ability of FCE.

\textbf{Comparison.} To verify the effectiveness and robustness of FCE for modeling text instances, we conduct comparisons with the recent SOTA arbitrated-shaped text detectors, TextRay~\cite{wang2020textray}. Results of Fig.~\ref{fig_fit-vs} show that TextRay fails to fit the ground-truth closely for highly-curved texts, while our FCE obtains accurate approximation. Note that our FCE uses $22$ dimensional parameters only while TextRay 44, which is 2 times as big as ours.


\subsection{Evaluation of FCENet}\label{sec_fcenent}

\textbf{Ablation Studies.}
To evaluate the effectiveness of the components of FCENet, we conducted ablation studies on both CTW1500 and Total-Text dataset, as shown in Table~\ref{table_abla}. The results indicate that the text center region (TCR) loss of the classification branch and the proposed regression loss (Eq.~\ref{eq-reloss}) of the regression branch can dramatically improve the performance of FCENet.

\begin{table}[t]
\centering
\small{
\caption{Ablation studies. ``TCR'' denotes text center region loss, ``RL'' denotes regression loss in Eq.~\ref{eq-reloss}.}
\begin{tabular}{|c|c||c|c|c||c|c|c|}
  \hline
      \multirow{2}{*}{TCR}& \multirow{2}{*}{RL}&
      \multicolumn{3}{c|}{CTW1500}&\multicolumn{3}{c|}{Total-Text}\cr\cline{3-8}
      &&R($\%$)      & P($\%$)      & F($\%$)& R($\%$)      & P($\%$)      & F($\%$)\\
  \hline
  -      & -     &   74.2   &   80.2   &   77.1   &71.2   &   81.6   &   76.1 \\
  -      &$\surd$&   78.8   &   83.8   &   81.3   &73.2   &   84.7   &   78.6 \\
  $\surd$& -     &   74.1   &   83.6   &   78.6   &72.0   &   81.6   &   76.5 \\
  \hline
  \hline
  $\surd$&$\surd$&   83.4   &   87.6   &   85.5   &82.5   &   89.3   &   85.8 \\
  \hline
\end{tabular}\label{table_abla}
}
\end{table}


\begin{table}[t]
\caption{{Generalization ability} evaluation on CTW1500 with different amounts of training data.} 
\centering
\small{
\begin{tabular}{|c|r|c|c|c|}
  \hline

      Data& Methods &R($\%$)      & P($\%$)      & F($\%$) \\
  \hline
  \hline
  \multirow{3}{*}{50$\%$}&
  DRRG~\cite{zhang2020deep}	             &61.1	&76.6	&68.0           \\
  &TextRay~\cite{wang2020textray}         &75.5  &77.8   &76.6          \\
  &ABCNet~\cite{liu2020abcnet}	         &71.1	&80.6	&75.6           \\
  &\textbf{FCENet\dag}   &\textbf{76.2}	&\textbf{84.9}	&\textbf{80.3}      \\
  \hline
  \hline
  \multirow{3}{*}{25$\%$}&
  DRRG~\cite{zhang2020deep}	             &44.5	&70.7	&54.7           \\
  &TextRay~\cite{wang2020textray}         &67.9  &74.9   &71.2          \\
  &ABCNet~\cite{liu2020abcnet}	         &70.0	&75.5	&72.7           \\
  &\textbf{FCENet\dag}	    &\textbf{75.7}	&\textbf{81.9}	&\textbf{78.7}  \\
  \hline
\end{tabular}\label{table_gen}}
\end{table}

\textbf{Generalization Ability.}
Benefiting from the FCE representation, FCENet requires the simple IFT and NMS post-processing only to reconstruct the complex text contours. Moreover, FCE can generate compact text representations, which
allows our FCENet has better generalization ability comparing to the SOTA methods. We made comparisons with DRRG~\cite{zhang2020deep}, TextRay~\cite{wang2020textray}, ABCNet~\cite{liu2020abcnet} and our proposed FCENet on CTW1500 dataset using different amounts of training data, as shown in Table~\ref{table_gen}.

The results show that the performance of the other methods would drop dramatically when training data is reduced to $50\%$ and $25\%$ of the original. In contrast, our FCENet maintains good accuracy, where all accuracy recall, precision and F-measure are over $73\%$ (precision maintains even over $80\%$). The results indicate the good generalization ability of our FCENet, and show the wide application potential of our method, especially in the practical scenarios with limited training samples.

\begin{figure*}[t]
  \centering
  \includegraphics[width=1.4in]{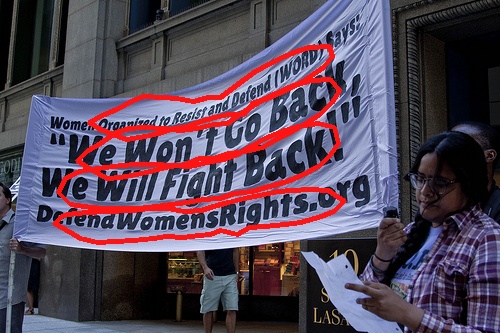}\hspace{1.5pt}
  \includegraphics[width=1.4in]{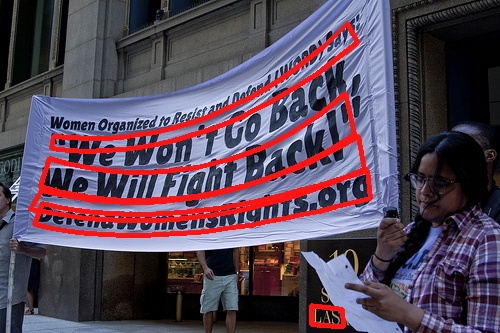}\hspace{1.5pt}
  \includegraphics[width=1.4in]{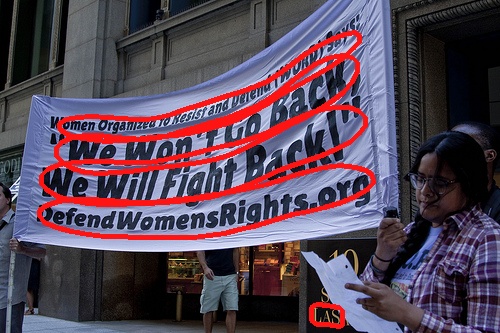}\hspace{1.5pt}
  \includegraphics[width=1.4in]{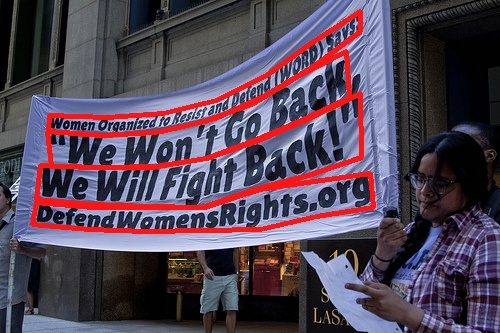}\vspace{3pt}\\

  \includegraphics[width=1.4in]{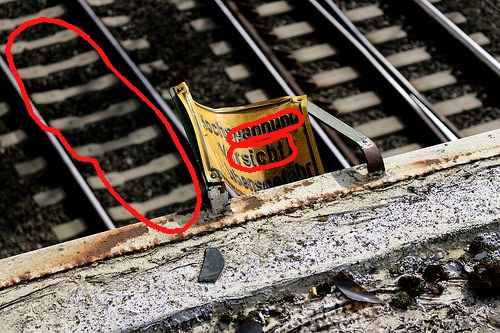}\hspace{1.5pt}
  \includegraphics[width=1.4in]{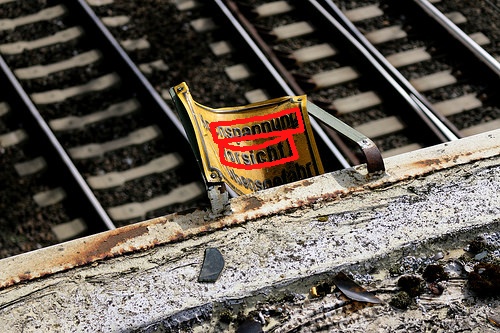}\hspace{1.5pt}
  \includegraphics[width=1.4in]{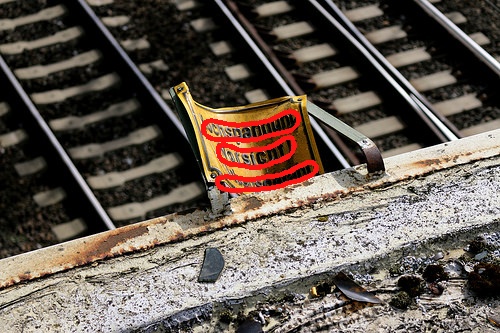}\hspace{1.5pt}
  \includegraphics[width=1.4in]{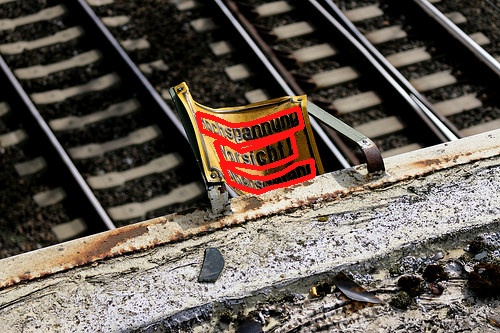}\vspace{-6pt}\\

  \subfloat[TextRay~\cite{wang2020textray}]{\includegraphics[width=1.4in]{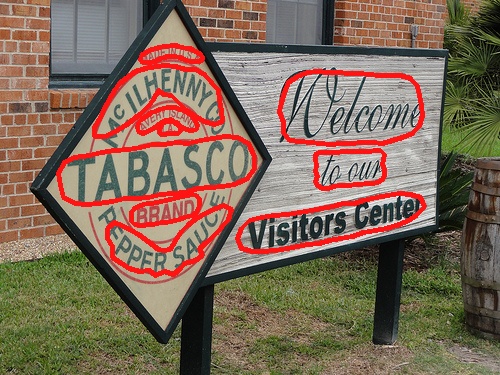}}\hspace{1.5pt}
  \subfloat[ABCNet~\cite{liu2020abcnet}]{\includegraphics[width=1.4in]{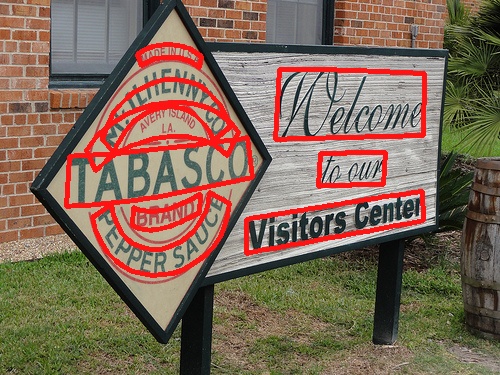}}\hspace{1.5pt}
  \subfloat[Ours]{\includegraphics[width=1.4in]{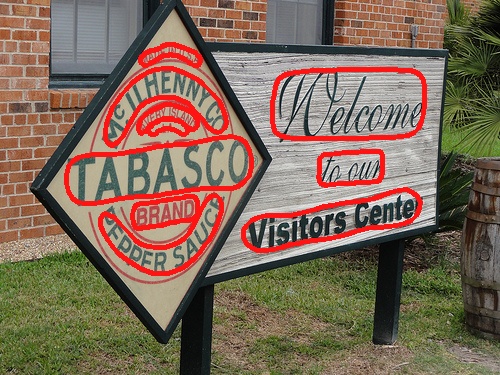}}\hspace{1.5pt}
  \subfloat[GT]{\includegraphics[width=1.4in]{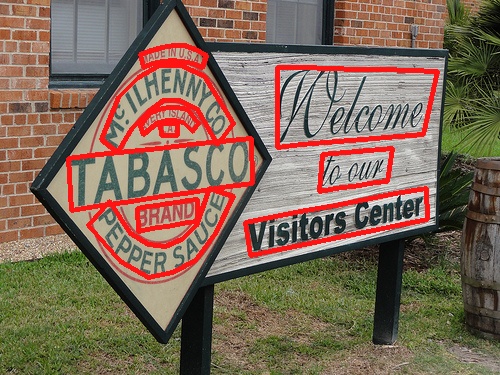}}
 \caption{Qualitative comparisons with TextRay~\cite{wang2020textray} and ABCNet~\cite{liu2020abcnet} on selected challenging samples in CTW1500 .}\label{fig_ch}
\end{figure*}

\subsection{Evaluation on Benchmark Datasets}
We made extensive comparison with most recent SOTA methods on different datasets, as shown in Table~\ref{table3}. The results illustrate that our FCENet obtains the best performance of precision (P) and F-measure (F), and achieves competitive performance of recall (R) on CTW1500 and Total-Text datasets of arbitrary-shaped texts. Note that most of the previous methods, except ContourNet~\cite{wang2020contournet}, require extra training data to obtain its best performance, but our FCENet is trained without it. On ICDAR15 dataset of multi-orientated and street-viewed texts, FCENet also achieves competitive results without additional setup.

Moreover, FCENet has simple network architecture, and efficient post-processing (\ie, IFT and NMS), which makes it easy to implement and very practical. Note that even through FCENEet$\dag$ uses ResNet50 as a backbone without DCN, which has the same backbone of ABCNet~\cite{liu2020abcnet}, it still obtain competitive results on different datasets in Table~\ref{table3}.



\begin{table}[t]
\caption{Quantitative comparison on the highly-curved text subset of CTW1500. 'Ext.' denotes extra training data.}
\centering
\begin{tabular}{|r|c|c|c|c|}
  \hline

  Methods                       & Ext.      &R($\%$)    &P($\%$)      &F($\%$) \\
  \hline
  \hline
  TextRay~\cite{wang2020textray}&$\surd$    &71.2	    &77.0	      &74.0\\
  \hline
  ABCNet~\cite{liu2020abcnet}   &$\surd$    &66.7       &79.8         &72.6\\
  \hline
  \textbf{FCENet}                 &-  &\textbf{74.7}	&\textbf{84.3}	&\textbf{79.2}\\
  \hline
\end{tabular}\label{table_ch}
\end{table}

\subsection{Evaluation on Highly-curved Subset}\label{sec_exp_chall}
Since CTW1500 still contains a large amount of non-curved texts, We selected highly-curved texts from it to build a challenging subset. Comparisons were made among most recent  methods with explicit text shape modeling~\cite{wang2020textray,liu2020abcnet}, \ie, TextRay~\cite{wang2020textray} and ABCNet~\cite{liu2020abcnet}.

To built the challenging subset, we discard the ``simple'' non-curved texts but reserve highly-curved texts. We utilize an algorithm to select the subset (total 106 samples), based on the observation that when we remove one of ground truth annotation points except the head and tail, the area of highly-curved text will change greatly. We computed the area of the annotation polygon before $A_{bef}$ and after $A_{aft}$ removing a point in ground truth annotations, and selected samples if $|A_{bef}-A_{aft}|/A_{bef}\geq0.07$.

Qualitative and quantitative comparisons were shown in Fig.~\ref{fig_ch} and Table~\ref{table_ch}, respectively. The results indicate that FCE is complementary to TextRay~\cite{wang2020textray} and ABCNet~\cite{liu2020abcnet} for explicitly modeling irregular text instances, and also show the effectiveness of FCENet for highly-curved text detection.

\section{Conclusion}
This paper focuses on the explicit shape modeling for arbitrary-shaped text detection. We propose Fourier contour embedding method, which allows to approximate any closed shapes accurately. Then, we propose FCENet which first predicts Fourier signature vectors of text instances in the Fourier domain, and then reconstructs text contour point sequences in the image spatial domain via the Inverse Fourier Transformation. FCENet can be optimized in an end-to-end manner, and be implemented without any complex post processing. Extensive evaluation were performed for the proposed FCE and FCENet.  Experimental results validate the representation capability of FCE, especially on highly-curved texts, and good generalization of FCENet when training with small samples. Moreover, it shows that FCENet achieves the SOTA performance on CTW1500, Total-Text and competitive results on ICDAR2015.

{\scriptsize
\bibliographystyle{ieee_fullname}
\bibliography{cvpr2021}
}

\end{document}